\PassOptionsToPackage{unicode}{hyperref}
\PassOptionsToPackage{hyphens}{url}
\PassOptionsToPackage{dvipsnames,svgnames,x11names}{xcolor}
\documentclass[
]{article}

\usepackage{amsmath,amssymb}
\usepackage{iftex}
\ifPDFTeX
  \usepackage[T1]{fontenc}
  \usepackage[utf8]{inputenc}
  \usepackage{textcomp} 
\else 
  \usepackage{unicode-math}
  \defaultfontfeatures{Scale=MatchLowercase}
  \defaultfontfeatures[\rmfamily]{Ligatures=TeX,Scale=1}
\fi
\usepackage{lmodern}
\ifPDFTeX\else  
\fi
\IfFileExists{upquote.sty}{\usepackage{upquote}}{}
\IfFileExists{microtype.sty}{
  \usepackage[]{microtype}
  \UseMicrotypeSet[protrusion]{basicmath} 
}{}
\makeatletter
\@ifundefined{KOMAClassName}{
  \IfFileExists{parskip.sty}{%
    \usepackage{parskip}
  }{
    \setlength{\parindent}{0pt}
    \setlength{\parskip}{6pt plus 2pt minus 1pt}}
}{
  \KOMAoptions{parskip=half}}
\makeatother
\usepackage{xcolor}
\ifx\paragraph\undefined\else
  \let\oldparagraph\paragraph
  \renewcommand{\paragraph}[1]{\oldparagraph{#1}\mbox{}}
\fi
\ifx\subparagraph\undefined\else
  \let\oldsubparagraph\subparagraph
  \renewcommand{\subparagraph}[1]{\oldsubparagraph{#1}\mbox{}}
\fi

\usepackage{color}
\usepackage{fancyvrb}

\DefineVerbatimEnvironment{Highlighting}{Verbatim}{commandchars=\\\{\}}
\usepackage{framed}
\definecolor{shadecolor}{RGB}{241,243,245}
\newenvironment{Shaded}{\begin{snugshade}}{\end{snugshade}}

\newcommand{\NormalTok}[1]{\textcolor[rgb]{0.00,0.23,0.31}{#1}}

\providecommand{\tightlist}{%
  \setlength{\itemsep}{0pt}\setlength{\parskip}{0pt}}\usepackage{longtable,booktabs,array}
\usepackage{calc} 
\usepackage{etoolbox}
\makeatletter
\patchcmd\longtable{\par}{\if@noskipsec\mbox{}\fi\par}{}{}
\makeatother
\IfFileExists{footnotehyper.sty}{\usepackage{footnotehyper}}{\usepackage{footnote}}
\makesavenoteenv{longtable}
\usepackage{graphicx}
\makeatletter
\def\maxwidth{\ifdim\Gin@nat@width>\linewidth\linewidth\else\Gin@nat@width\fi}
\def\maxheight{\ifdim\Gin@nat@height>\textheight\textheight\else\Gin@nat@height\fi}
\makeatother
\setkeys{Gin}{width=\maxwidth,height=\maxheight,keepaspectratio}
\makeatletter
\def\fps@figure{htbp}
\makeatother
\NewDocumentCommand\citeproctext{}{}

\makeatletter
 \let\@cite@ofmt\@firstofone
 \def\@biblabel#1{}
 \def\@cite#1#2{{#1\if@tempswa , #2\fi}}
\makeatother
\newlength{\cslhangindent}
\newlength{\csllabelwidth}
\newenvironment{CSLReferences}[2] 
 {\begin{list}{}{%
  \setlength{\itemindent}{0pt}
  \setlength{\leftmargin}{0pt}
  \setlength{\parsep}{0pt}
  \ifodd #1
   \setlength{\leftmargin}{\cslhangindent}
   \setlength{\itemindent}{-1\cslhangindent}
  \fi
  \setlength{\itemsep}{#2\baselineskip}}}
 {\end{list}}
\usepackage{calc}

\usepackage{arxiv}
\usepackage{orcidlink}
\usepackage{amsmath}
\usepackage[T1]{fontenc}
\makeatletter
\@ifpackageloaded{tcolorbox}{}{\usepackage[skins,breakable]{tcolorbox}}
\@ifpackageloaded{fontawesome5}{}{\usepackage{fontawesome5}}
\definecolor{quarto-callout-color}{HTML}{909090}
\definecolor{quarto-callout-note-color}{HTML}{0758E5}
\definecolor{quarto-callout-important-color}{HTML}{CC1914}
\definecolor{quarto-callout-warning-color}{HTML}{EB9113}
\definecolor{quarto-callout-tip-color}{HTML}{00A047}
\definecolor{quarto-callout-caution-color}{HTML}{FC5300}
\definecolor{quarto-callout-color-frame}{HTML}{acacac}
\definecolor{quarto-callout-note-color-frame}{HTML}{4582ec}
\definecolor{quarto-callout-important-color-frame}{HTML}{d9534f}
\definecolor{quarto-callout-warning-color-frame}{HTML}{f0ad4e}
\definecolor{quarto-callout-tip-color-frame}{HTML}{02b875}
\definecolor{quarto-callout-caution-color-frame}{HTML}{fd7e14}
\makeatother
\makeatletter
\@ifpackageloaded{caption}{}{\usepackage{caption}}
\AtBeginDocument{%
\ifdefined\contentsname
  \renewcommand*\contentsname{Table of contents}
\else
  \newcommand\contentsname{Table of contents}
\fi
\ifdefined\listfigurename
  \renewcommand*\listfigurename{List of Figures}
\else
  \newcommand\listfigurename{List of Figures}
\fi
\ifdefined\listtablename
  \renewcommand*\listtablename{List of Tables}
\else
  \newcommand\listtablename{List of Tables}
\fi
\ifdefined\figurename
  \renewcommand*\figurename{Figure}
\else
  \newcommand\figurename{Figure}
\fi
\ifdefined\tablename
  \renewcommand*\tablename{Table}
\else
  \newcommand\tablename{Table}
\fi
}
\@ifpackageloaded{float}{}{\usepackage{float}}
\floatstyle{ruled}
\@ifundefined{c@chapter}{\newfloat{codelisting}{h}{lop}}{\newfloat{codelisting}{h}{lop}[chapter]}
\floatname{codelisting}{Listing}

\makeatother
\makeatletter
\@ifpackageloaded{caption}{}{\usepackage{caption}}
\@ifpackageloaded{subcaption}{}{\usepackage{subcaption}}
\makeatother
\ifLuaTeX
  \usepackage{selnolig}  
\fi
\usepackage{bookmark}

\IfFileExists{xurl.sty}{\usepackage{xurl}}{} 
\hypersetup{
  pdftitle={Simplifying Hyperparameter Tuning in Online Machine Learning---The spotRiverGUI},
  pdfauthor={Thomas Bartz-Beielstein},
  pdfkeywords={hyperparameter tuning, online machine learning, streaming
data, river, sequential parameter optimization},
  colorlinks=true,
  linkcolor={blue},
  filecolor={Maroon},
  citecolor={Blue},
  urlcolor={Blue},
  pdfcreator={LaTeX via pandoc}}

\renewcommand{\today}{Feb 18, 2024}
\newcommand{\runninghead}{A Preprint }
\renewcommand{\runninghead}{A Preprint }
\title{Simplifying Hyperparameter Tuning in Online Machine
Learning---The spotRiverGUI}
\author{\textbf{Thomas
Bartz-Beielstein}~\orcidlink{0000-0002-5938-5158}\\\\Bartz \& Bartz
GmbH\\51643 Gummersbach,
Germany\\\href{mailto:bartzbeielstein@gmail.com}{bartzbeielstein@gmail.com}}
\date{Feb 18, 2024}
\begin{document}
\maketitle
\begin{abstract}
Batch Machine Learning (BML) reaches its limits when dealing with very
large amounts of streaming data. This is especially true for available
memory, handling drift in data streams, and processing new, unknown
data. Online Machine Learning (OML) is an alternative to BML that
overcomes the limitations of BML. OML is able to process data in a
sequential manner, which is especially useful for data streams. The
\texttt{river} package is a Python OML-library, which provides a variety
of online learning algorithms for classification, regression,
clustering, anomaly detection, and more. The \texttt{spotRiver} package
provides a framework for hyperparameter tuning of OML models. The
\texttt{spotRiverGUI} is a graphical user interface for the
\texttt{spotRiver} package. The \texttt{spotRiverGUI} releases the user
from the burden of manually searching for the optimal hyperparameter
setting. After the data is provided, users can compare different OML
algorithms from the powerful \texttt{river} package in a convenient way
and tune the selected algorithms very efficiently.
\end{abstract}
{\bfseries \emph Keywords}
\def\sep{\textbullet\ }
hyperparameter tuning \sep online machine learning \sep streaming
data \sep river \sep 
sequential parameter optimization

\section{Introduction}\label{introduction}

Batch Machine Learning (BML) often encounters limitations when
processing substantial volumes of streaming data (Keller-McNulty 2004;
Gaber, Zaslavsky, and Krishnaswamy 2005; Aggarwal 2007). These
limitations become particularly evident in terms of available memory,
managing drift in data streams (Bifet and Gavaldà 2007, 2009; Gama et
al. 2004; Bartz-Beielstein 2024c), and processing novel, unclassified
data (Bifet 2010), (Dredze, Oates, and Piatko 2010). As a solution,
Online Machine Learning (OML) serves as an effective alternative to BML,
adeptly addressing these constraints. OML's ability to sequentially
process data proves especially beneficial for handling data streams
(Bifet et al. 2010a; Masud et al. 2011; Gama, Sebastião, and Rodrigues
2013; Putatunda 2021; Bartz-Beielstein and Hans 2024).

The Online Machine Learning (OML) methods provided by software packages
such as \texttt{river} (Montiel et al. 2021) or \texttt{MOA} (Bifet et
al. 2010b) require the specification of many hyperparameters. To give an
example, Hoeffding trees (Hoeglinger and Pears 2007), which are very
popular in OML, offer a variety of ``splitters'' to generate subtrees.
There are also several methods to limit the tree size, ensuring time and
memory requirements remain manageable. Given the multitude of
parameters, manually searching for the optimal hyperparameter setting
can be a daunting and often futile task due to the complexity of
possible combinations. This article elucidates how automatic
hyperparameter optimization, or ``tuning'', can be achieved. Beyond
optimizing the OML process, Hyperparameter Tuning (HPT) executed with
the Sequential Parameter Optimization Toolbox (SPOT) enhances the
explainability and interpretability of OML procedures. This can result
in a more efficient, resource-conserving algorithm, contributing to the
concept of ``Green AI''.

\begin{tcolorbox}[enhanced jigsaw, title=\textcolor{quarto-callout-note-color}{\faInfo}\hspace{0.5em}{Note}, left=2mm, colframe=quarto-callout-note-color-frame, opacitybacktitle=0.6, opacityback=0, toptitle=1mm, bottomtitle=1mm, rightrule=.15mm, leftrule=.75mm, coltitle=black, breakable, toprule=.15mm, bottomrule=.15mm, colbacktitle=quarto-callout-note-color!10!white, arc=.35mm, titlerule=0mm, colback=white]

Note: This document refers to \texttt{spotRiverGUI} version 0.0.26 which
was released on Feb 18, 2024 on GitHub, see:
\url{https://github.com/sequential-parameter-optimization/spotGUI/tree/main}.
The GUI is under active development and new features will be added soon.

\end{tcolorbox}

This article describes the \texttt{spotRiverGUI}, which is a graphical
user interface for the \texttt{spotRiver} package. The GUI allows the
user to select the task, the data set, the preprocessing model, the
metric, and the online machine learning model. The user can specify the
experiment duration, the initial design, and the evaluation options. The
GUI provides information about the data set and allows the user to save
and load experiments. It also starts and stops a tensorboard process to
observe the tuning online and provides an analysis of the hyperparameter
tuning process. The \texttt{spotRiverGUI} releases the user from the
burden of manually searching for the optimal hyperparameter setting.
After providing the data, users can compare different OML algorithms
from the powerful \texttt{river} package in a convenient way and tune
the selected algorithm very efficiently.

This article is structured as follows:

Section~\ref{sec-starting-gui} describes how to install the software. It
also explains how the \texttt{spotRiverGUI} can be started.
Section~\ref{sec-binary-classification} describes the binary
classification task and the options available in the
\texttt{spotRiverGUI}. Section~\ref{sec-regression} provides information
about the planned regression task. Section~\ref{sec-showing-data}
describes how the data can be visualized in the \texttt{spotRiverGUI}.
Section~\ref{sec-saving-loading} provides information about saving and
loading experiments. Section~\ref{sec-running-experiment} describes how
to start an experiment and how the associated tensorboard process can be
started and stopped. Section~\ref{sec-analysis} provides information
about the analysis of the results from the hyperparameter tuning
process. Section~\ref{sec-summary} concludes the article and provides an
outlook.

\section{Installation and Starting}\label{sec-starting-gui}

\subsection{Installation}\label{installation}

We strongly recommend using a virtual environment for the installation
of the \texttt{river}, \texttt{spotRiver}, \texttt{build} and
\texttt{spotRiverGUI} packages.

Miniforge, which holds the minimal installers for Conda, is a good
starting point. Please follow the instructions on
\url{https://github.com/conda-forge/miniforge}. Using Conda, the
following commands can be used to create a virtual environment (Python
3.11 is recommended):

\begin{Shaded}
\begin{Highlighting}[]
\NormalTok{\textgreater{}\textgreater{} conda create {-}n myenv python=3.11}
\NormalTok{\textgreater{}\textgreater{} conda activate myenv}
\end{Highlighting}
\end{Shaded}

Now the \texttt{river} and \texttt{spotRiver} packages can be installed:

\begin{Shaded}
\begin{Highlighting}[]
\NormalTok{\textgreater{}\textgreater{} (myenv) pip install river spotRiver build}
\end{Highlighting}
\end{Shaded}

Although the \texttt{spotGUI} package is available on PyPI, we recommend
an installation from the GitHub repository
\href{./https://github.com/sequential-parameter-optimization/spotGUI}{https://github.com/sequential-parameter-optimization/spotGUI},
because the \texttt{spotGUI} package is under active development and new
features will be added soon. The installation from the GitHub repository
is done by executing the following command:

\begin{Shaded}
\begin{Highlighting}[]
\NormalTok{\textgreater{}\textgreater{} (myenv) git clone git@github.com:sequential{-}parameter{-}optimization/spotGUI.git}
\end{Highlighting}
\end{Shaded}

Building the \texttt{spotGUI} package is done by executing the following
command:

\begin{Shaded}
\begin{Highlighting}[]
\NormalTok{\textgreater{}\textgreater{} (myenv) cd spotGUI}
\NormalTok{\textgreater{}\textgreater{} (myenv) python {-}m build}
\end{Highlighting}
\end{Shaded}

Now the \texttt{spotRiverGUI} package can be installed:

\begin{Shaded}
\begin{Highlighting}[]
\NormalTok{\textgreater{}\textgreater{} (myenv) pip install dist/spotGUI{-}0.0.26.tar.gz}
\end{Highlighting}
\end{Shaded}

\subsection{Starting the GUI}\label{starting-the-gui}

The GUI can be started by executing the \texttt{spotRiverGUI.py} file in
the \texttt{spotGUI/spotRiverGUI} directory. Change to the
\texttt{spotRiverGUI} directory and start the GUI:

\begin{Shaded}
\begin{Highlighting}[]
\NormalTok{\textgreater{}\textgreater{} (myenv) cd spotGUI/spotRiverGUI}
\NormalTok{\textgreater{}\textgreater{} (myenv) python spotRiverGUI.py}
\end{Highlighting}
\end{Shaded}

\begin{figure}

\centering{

\includegraphics{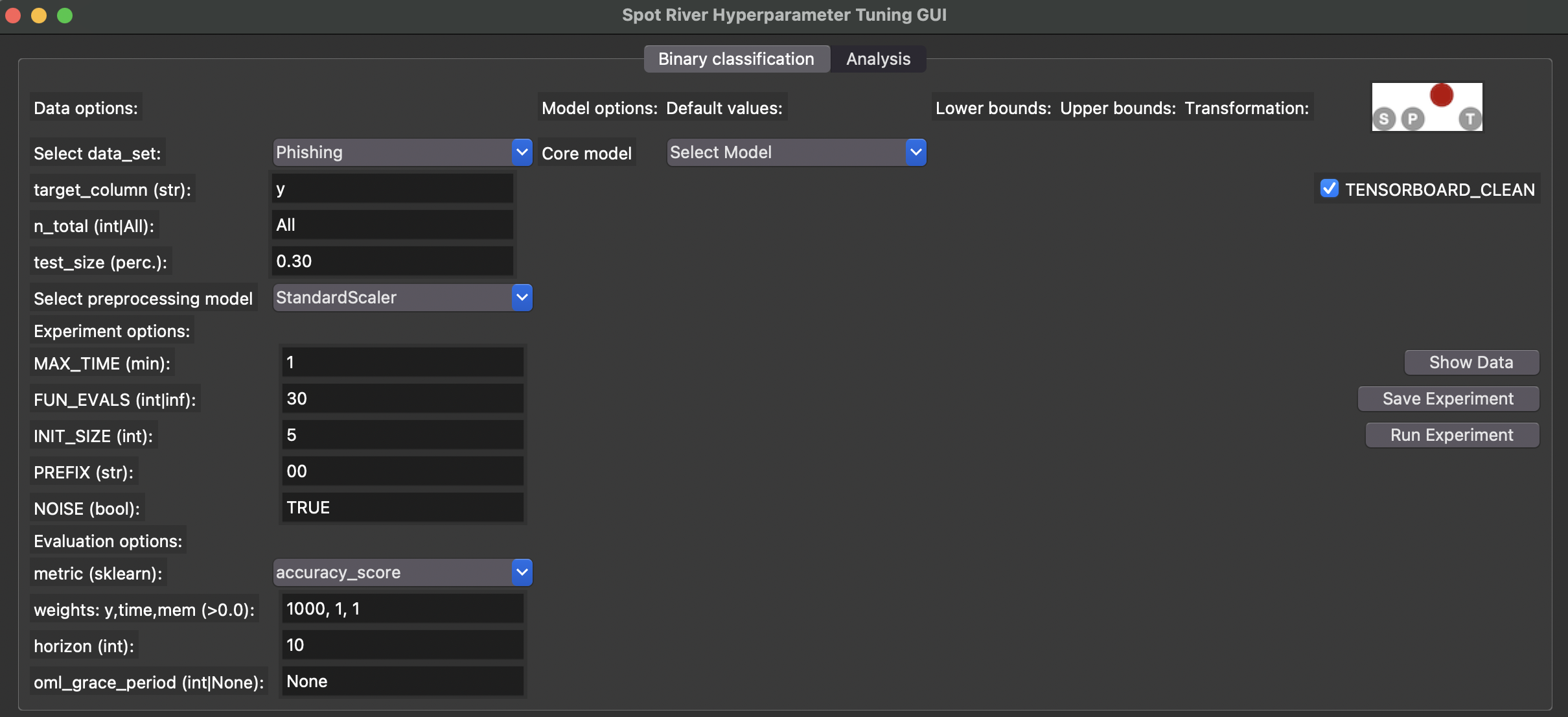}

}

\caption{\label{fig-spotRiverGUI-00}spotriver GUI}

\end{figure}

The GUI window will open, as shown in Figure~\ref{fig-spotRiverGUI-00}.

After the GUI window has opened, the user can select the task.
Currently, \texttt{Binary\ Classification} is available. Further tasks
like \texttt{Regression} will be available soon.

Depending on the task, the user can select the data set, the
preprocessing model, the metric, and the online machine learning model.

\section{Binary Classification}\label{sec-binary-classification}

\subsection{Binary Classification
Options}\label{binary-classification-options}

If the \texttt{Binary\ Classification} task is selected, the user can
select pre-specified data sets from the \texttt{Data} drop-down menu.

\subsubsection{River Data Sets}\label{sec-river-datasets}

The following data sets from the \texttt{river} package are available
(the descriptions are taken from the \texttt{river} package):

\begin{itemize}
\tightlist
\item
  \texttt{Bananas}: An artificial dataset where instances belongs to
  several clusters with a banana shape.There are two attributes that
  correspond to the x and y axis, respectively. More:
  \url{https://riverml.xyz/dev/api/datasets/Bananas/}.
\item
  \texttt{CreditCard}: Credit card frauds. The datasets contains
  transactions made by credit cards in September 2013 by European
  cardholders. Feature `\texttt{Class}' is the response variable and it
  takes value 1 in case of fraud and 0 otherwise. More:
  \url{https://riverml.xyz/dev/api/datasets/CreditCard/}.
\item
  \texttt{Elec2}: Electricity prices in New South Wales. This is a
  binary classification task, where the goal is to predict if the price
  of electricity will go up or down. This data was collected from the
  Australian New South Wales Electricity Market. In this market, prices
  are not fixed and are affected by demand and supply of the market.
  They are set every five minutes. Electricity transfers to/from the
  neighboring state of Victoria were done to alleviate fluctuations.
  More: \url{https://riverml.xyz/dev/api/datasets/Elec2/}.
\item
  \texttt{Higgs}: The data has been produced using Monte Carlo
  simulations. The first 21 features (columns 2-22) are kinematic
  properties measured by the particle detectors in the accelerator. The
  last seven features are functions of the first 21 features; these are
  high-level features derived by physicists to help discriminate between
  the two classes. More:
  \url{https://riverml.xyz/dev/api/datasets/Higgs/}.
\item
  \texttt{HTTP}: HTTP dataset of the KDD 1999 cup. The goal is to
  predict whether or not an HTTP connection is anomalous or not. The
  dataset only contains 2,211 (0.4\%) positive labels. More:
  \url{https://riverml.xyz/dev/api/datasets/HTTP/}.
\item
  \texttt{Phishing}: Phishing websites. This dataset contains features
  from web pages that are classified as phishing or
  not.\url{https://riverml.xyz/dev/api/datasets/Phishing/}
\end{itemize}

\subsubsection{User Data Sets}\label{user-data-sets}

Besides the \texttt{river} data sets described in
Section~\ref{sec-river-datasets}, the user can also select a
user-defined data set. Currently, comma-separated values (CSV) files are
supported. Further formats will be supported soon. The user-defined CSV
data set must be a binary classification task with the target variable
in the last column. The first row must contain the column names. If the
file is copied to the subdirectory \texttt{userData}, the user can
select the data set from the \texttt{Data} drop-down menu.

As an example, we have provided a CSV-version of the \texttt{Phishing}
data set. The file is located in the \texttt{userData} subdirectory and
is called \texttt{PhishingData.csv}. It contains the columns
\texttt{empty\_server\_form\_handler}, \texttt{popup\_window},
\texttt{https}, \texttt{request\_from\_other\_domain},
\texttt{anchor\_from\_other\_domain}, \texttt{is\_popular},
\texttt{long\_url}, \texttt{age\_of\_domain}, \texttt{ip\_in\_url}, and
\texttt{is\_phishing}. The first few lines of the file are shown below
(modified due to formatting reasons):

\begin{Shaded}
\begin{Highlighting}[]
\NormalTok{empty\_server\_form\_handler,...,is\_phishing}
\NormalTok{0.0,0.0,0.0,0.0,0.0,0.5,1.0,1,1,1}
\NormalTok{1.0,0.0,0.5,0.5,0.0,0.5,0.0,1,0,1}
\NormalTok{0.0,0.0,1.0,0.0,0.5,0.5,0.0,1,0,1}
\NormalTok{0.0,0.0,1.0,0.0,0.0,1.0,0.5,0,0,1}
\end{Highlighting}
\end{Shaded}

Based on the required format, we can see that \texttt{is\_phishing} is
the target column, because it is the last column of the data set.

\subsubsection{Stream Data Sets}\label{stream-data-sets}

Forthcoming versions of the GUI will support stream data sets, e.g, the
Friedman-Drift generator (Ikonomovska 2012) or the SEA-Drift generator
(Street and Kim 2001). The Friedman-Drift generator was also used in the
hyperparameter tuning study in Bartz-Beielstein (2024b).

\subsubsection{Data Set Options}\label{data-set-options}

Currently, the user can select the following parameters for the data
sets:

\begin{itemize}
\tightlist
\item
  \texttt{n\_total}: The total number of instances. Since some data sets
  are quite large, the user can select a subset of the data set by
  specifying the \texttt{n\_total} value.
\item
  \texttt{test\_size}: The size of the test set in percent
  (\texttt{0.0\ -\ 1.0}). The training set will be
  \texttt{1.0\ -\ test\_size}.
\end{itemize}

The target column should be the last column of the data set. Future
versions of the GUI will support the selection of the
\texttt{target\_column} from the GUI. Currently, the value from the
field \texttt{target\_column} has not effect.

To compare different data scaling methods, the user can select the
preprocessing model from the \texttt{Preprocessing} drop-down menu.
Currently, the following preprocessing models are available:

\begin{itemize}
\tightlist
\item
  \texttt{StandardScaler}: Standardize features by removing the mean and
  scaling to unit variance.
\item
  \texttt{MinMaxScaler}: Scale features to a range.
\item
  \texttt{None}: No scaling is performed.
\end{itemize}

The \texttt{spotRiverGUI} will not provide sophisticated data
preprocessing methods. We assume that the data was preprocessed before
it is copied into the \texttt{userData} subdirectory.

\subsection{Experiment Options}\label{experiment-options}

Currently, the user can select the following options for specifying the
experiment duration:

\begin{itemize}
\tightlist
\item
  \texttt{MAX\_TIME}: The maximum time in minutes for the experiment.
\item
  \texttt{FUN\_EVALS}: The number of function evaluations for the
  experiment. This is the number of OML-models that are built and
  evaluated.
\end{itemize}

If the \texttt{MAX\_TIME} is reached or \texttt{FUN\_EVALS} OML models
are evaluated, the experiment will be stopped.

\begin{tcolorbox}[enhanced jigsaw, title=\textcolor{quarto-callout-note-color}{\faInfo}\hspace{0.5em}{Initial design is always evaluated}, left=2mm, colframe=quarto-callout-note-color-frame, opacitybacktitle=0.6, opacityback=0, toptitle=1mm, bottomtitle=1mm, rightrule=.15mm, leftrule=.75mm, coltitle=black, breakable, toprule=.15mm, bottomrule=.15mm, colbacktitle=quarto-callout-note-color!10!white, arc=.35mm, titlerule=0mm, colback=white]

\begin{itemize}
\tightlist
\item
  The initial design will always be evaluated before one of the stopping
  criteria is reached.
\item
  If the initial design is very large or the model evaluations are very
  time-consuming, the runtime will be larger than the \texttt{MAX\_TIME}
  value.
\end{itemize}

\end{tcolorbox}

Based on the \texttt{INIT\_SIZE}, the number of hyperparameter
configurations for the initial design can be specified. The initial
design is evaluated before the first surrogate model is built. A
detailed description of the initial design and the surrogate model based
hyperparameter tuning can be found in Bartz-Beielstein (2024a) and in
Bartz-Beielstein and Zaefferer (2022). The \texttt{spotPython} package
is used for the hyperparameter tuning process. It implements a robust
surrogate model based optimization method (Forrester, Sóbester, and
Keane 2008).

The \texttt{PREFIX} parameter can be used to specify the experiment
name.

The \texttt{spotPython} hyperparameter tuning program allows the user to
specify several options for the hyperparameter tuning process. The
\texttt{spotRiverGUI} will support more options in future versions.
Currently, the user can specify whether the outcome from the experiment
is noisy or deterministic. The corresponding parameter is called
\texttt{NOISE}. The reader is referred to Bartz-Beielstein (2024b) and
to the chapter ``Handling Noise''
(\url{https://sequential-parameter-optimization.github.io/Hyperparameter-Tuning-Cookbook/013_num_spot_noisy.html})
for further information about the \texttt{NOISE} parameter.

\subsection{Evaluation Options}\label{evaluation-options}

The user can select one of the following evaluation metrics for binary
classification tasks from the \texttt{metric} drop-down menu:

\begin{itemize}
\tightlist
\item
  \texttt{accuracy\_score}
\item
  \texttt{cohen\_kappa\_score}
\item
  \texttt{f1\_score}
\item
  \texttt{hamming\_loss}
\item
  \texttt{hinge\_loss}
\item
  \texttt{jaccard\_score}
\item
  \texttt{matthews\_corrcoef}
\item
  \texttt{precision\_score}
\item
  \texttt{recall\_score}
\item
  \texttt{roc\_auc\_score}
\item
  \texttt{zero\_one\_loss}
\end{itemize}

These metrics are based on the \texttt{scikit-learn} module (Pedregosa
et al. 2011), which implements several loss, score, and utility
functions to measure classification performance, see
\url{https://scikit-learn.org/stable/modules/model_evaluation.html\#classification-metrics}.
\texttt{spotRiverGUI} supports metrics that are computed from the
\texttt{y\_pred} and the \texttt{y\_true} values. The \texttt{y\_pred}
values are the predicted target values, and the \texttt{y\_true} values
are the true target values. The \texttt{y\_pred} values are generated by
the online machine learning model, and the \texttt{y\_true} values are
the true target values from the data set.

\begin{tcolorbox}[enhanced jigsaw, title=\textcolor{quarto-callout-note-color}{\faInfo}\hspace{0.5em}{Evaluation Metrics: Minimization and Maximization}, left=2mm, colframe=quarto-callout-note-color-frame, opacitybacktitle=0.6, opacityback=0, toptitle=1mm, bottomtitle=1mm, rightrule=.15mm, leftrule=.75mm, coltitle=black, breakable, toprule=.15mm, bottomrule=.15mm, colbacktitle=quarto-callout-note-color!10!white, arc=.35mm, titlerule=0mm, colback=white]

\begin{itemize}
\tightlist
\item
  Some metrics are minimized, and some are maximized. The
  \texttt{spotRiverGUI} will support the user in selecting the correct
  metric based on the task. For example, the \texttt{accuracy\_score} is
  maximized, and the \texttt{hamming\_loss} is minimized. The user can
  select the metric and \texttt{spotRiverGUI} will automatically
  determine whether the metric is minimized or maximized.
\end{itemize}

\end{tcolorbox}

In addition to the evaluation metric results, \texttt{spotRiver}
considers the time and memory consumption of the online machine learning
model. The \texttt{spotRiverGUI} will support the user in selecting the
time and memory consumption as additional evaluation metrics. By
modifying the weight vector, which is shown in the
\texttt{weights:\ y,\ time,\ mem} field, the user can specify the
importance of the evaluation metrics. For example, the weight vector
\texttt{1,0,0} specifies that only the \texttt{y} metric (e.g.,
accuracy) is considered. The weight vector \texttt{0,1,0} specifies that
only the time metric is considered. The weight vector \texttt{0,0,1}
specifies that only the memory metric is considered. The weight vector
\texttt{1,1,1} specifies that all metrics are considered. Any real
values (also negative ones) are allowed for the weights.

\begin{tcolorbox}[enhanced jigsaw, title=\textcolor{quarto-callout-note-color}{\faInfo}\hspace{0.5em}{The weight vector}, left=2mm, colframe=quarto-callout-note-color-frame, opacitybacktitle=0.6, opacityback=0, toptitle=1mm, bottomtitle=1mm, rightrule=.15mm, leftrule=.75mm, coltitle=black, breakable, toprule=.15mm, bottomrule=.15mm, colbacktitle=quarto-callout-note-color!10!white, arc=.35mm, titlerule=0mm, colback=white]

\begin{itemize}
\tightlist
\item
  The specification of adequate weights is highly problem dependent.
\item
  There is no generic setting that fits to all problems.
\end{itemize}

\end{tcolorbox}

As described in Bartz-Beielstein (2024a), a prediction horizon is used
for the comparison of the online-machine learning algorithms. The
\texttt{horizon} can be specified in the \texttt{spotRiverGUI} by the
user and is highly problem dependent. The \texttt{spotRiverGUI} uses the
\texttt{eval\_oml\_horizon} method from the \texttt{spotRiver} package,
which evaluates the online-machine learning model on a rolling horizon
basis.

In addition to the \texttt{horizon} value, the user can specify the
\texttt{oml\_grace\_period} value. During the
\texttt{oml\_grace\_period}, the OML-model is trained on the (small)
training data set. No predictions are made during this initial training
phase, but the memory and computation time are measured. Then, the
OML-model is evaluated on the test data set using a given (sklearn)
evaluation metric. The default value of the \texttt{oml\_grace\_period}
is \texttt{horizon}. For convenience, the value \texttt{horizon} is also
selected when the user specifies the \texttt{oml\_grace\_period} value
as \texttt{None}.

\begin{tcolorbox}[enhanced jigsaw, title=\textcolor{quarto-callout-note-color}{\faInfo}\hspace{0.5em}{The oml\_grace\_period}, left=2mm, colframe=quarto-callout-note-color-frame, opacitybacktitle=0.6, opacityback=0, toptitle=1mm, bottomtitle=1mm, rightrule=.15mm, leftrule=.75mm, coltitle=black, breakable, toprule=.15mm, bottomrule=.15mm, colbacktitle=quarto-callout-note-color!10!white, arc=.35mm, titlerule=0mm, colback=white]

\begin{itemize}
\tightlist
\item
  If the \texttt{oml\_grace\_period} is set to the size of the training
  data set, the OML-model is trained on the entire training data set and
  then evaluated on the test data set using a given (sklearn) evaluation
  metric.
\item
  This setting might be ``unfair'' in some cases, because the OML-model
  should learn online and not on the entire training data set.
\item
  Therefore, a small data set is recommended for the
  \texttt{oml\_grace\_period} setting and the prediction
  \texttt{horizon} is a recommended value for the
  \texttt{oml\_grace\_period} setting. The reader is referred to
  Bartz-Beielstein (2024a) for further information about the
  \texttt{oml\_grace\_period} setting.
\end{itemize}

\end{tcolorbox}

\subsection{Online Machine Learning Model
Options}\label{online-machine-learning-model-options}

The user can select one of the following online machine learning models
from the \texttt{coremodel} drop-down menu:

\begin{itemize}
\tightlist
\item
  \texttt{forest.AMFClassifier}: Aggregated Mondrian Forest classifier
  for online learning (Mourtada, Gaiffas, and Scornet 2019). This
  implementation is truly online, in the sense that a single pass is
  performed, and that predictions can be produced anytime. More:
  \url{https://riverml.xyz/dev/api/forest/AMFClassifier/}.
\item
  \texttt{tree.ExtremelyFastDecisionTreeClassifier}: Extremely Fast
  Decision Tree (EFDT) classifier (Manapragada, Webb, and Salehi 2018).
  Also referred to as the Hoeffding AnyTime Tree (HATT) classifier. In
  practice, despite the name, EFDTs are typically slower than a vanilla
  Hoeffding Tree to process data. More:
  \url{https://riverml.xyz/dev/api/tree/ExtremelyFastDecisionTreeClassifier/}.
\item
  \texttt{tree.HoeffdingTreeClassifier}: Hoeffding Tree or Very Fast
  Decision Tree classifier (Bifet et al. 2010a; Domingos and Hulten
  2000). More:
  \url{https://riverml.xyz/dev/api/tree/HoeffdingTreeClassifier/}.
\item
  \texttt{tree.HoeffdingAdaptiveTreeClassifier}: Hoeffding Adaptive Tree
  classifier (Bifet and Gavaldà 2009). More:
  \url{https://riverml.xyz/dev/api/tree/HoeffdingAdaptiveTreeClassifier/}.
\item
  \texttt{linear\_model.LogisticRegression}: Logistic regression
  classifier. More:
  \href{https://riverml.xyz/dev/api/linear-model/LogisticRegression/}{hhttps://riverml.xyz/dev/api/linear-model/LogisticRegression/}.
\end{itemize}

\begin{figure}

\centering{

\includegraphics{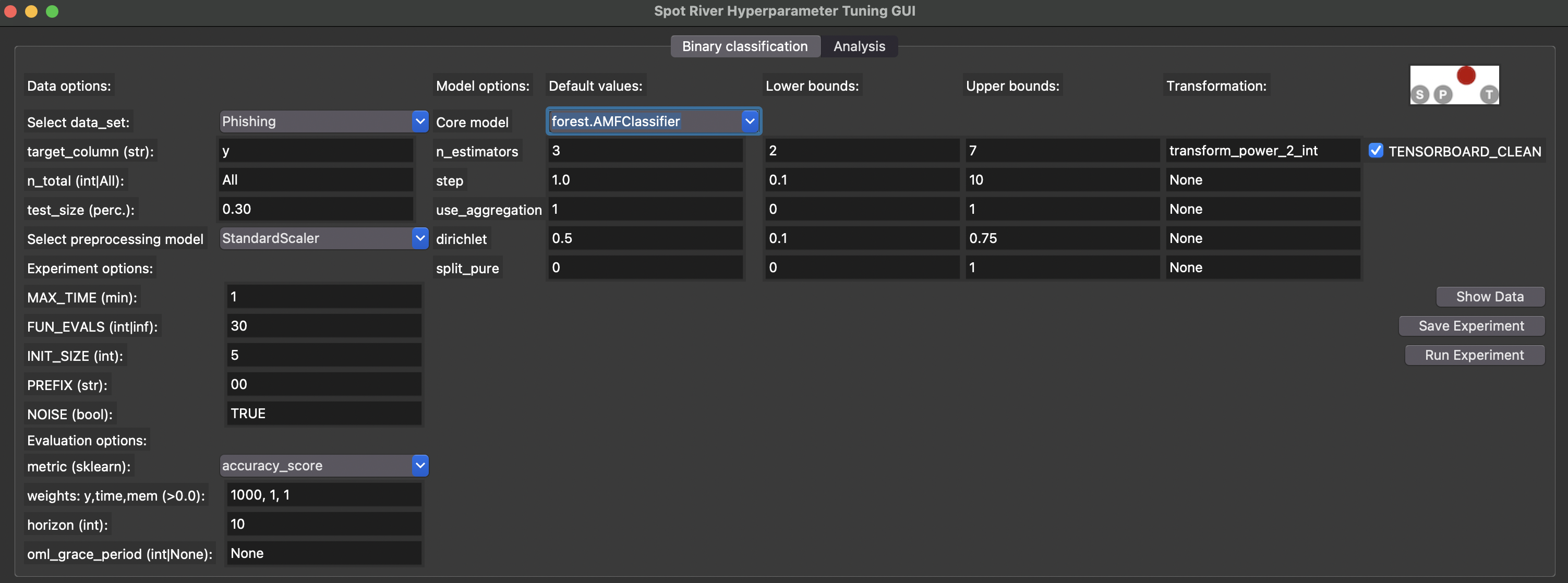}

}

\caption{\label{fig-spotRiverGUI-01}\texttt{spotRiverGUI} when
\texttt{forest.AMFClassifier} is selected}

\end{figure}

\begin{figure}

\centering{

\includegraphics{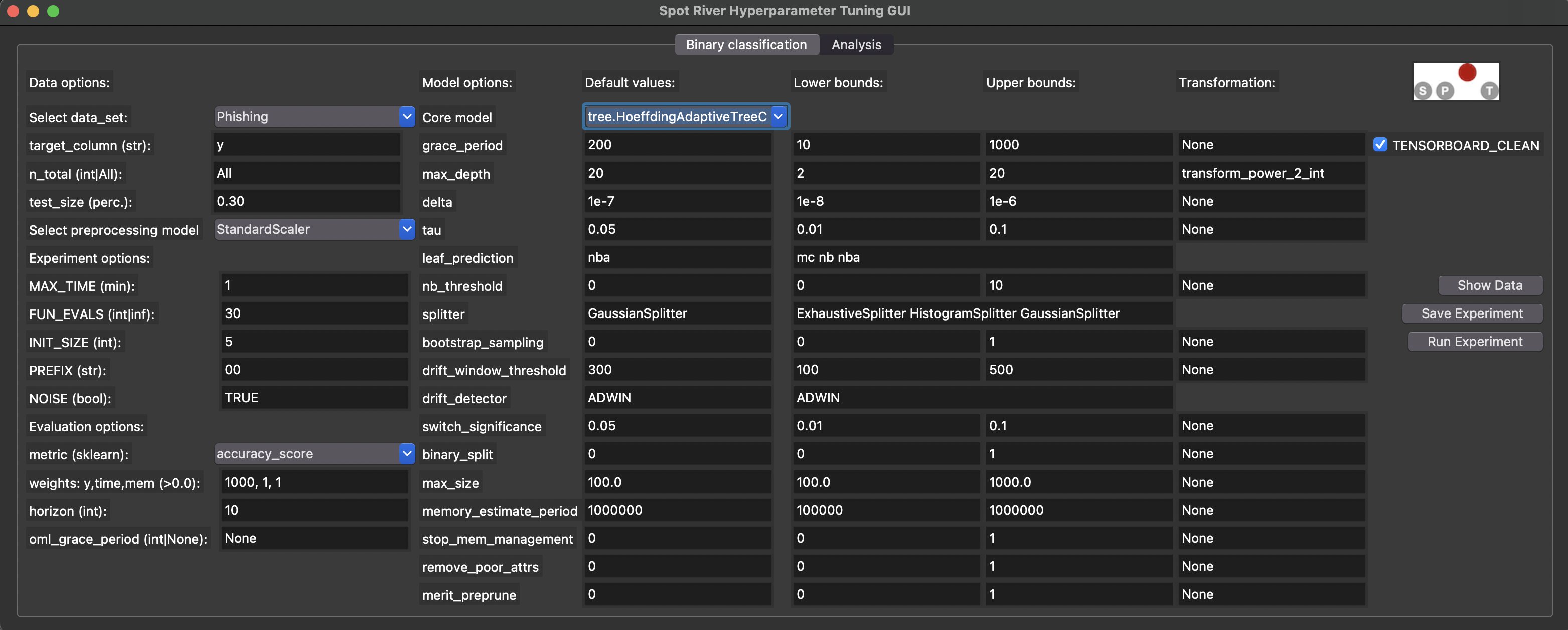}

}

\caption{\label{fig-spotRiverGUI-02}\texttt{spotRiverGUI} when
\texttt{tree.HoeffdingAdaptiveTreeClassifier} is selected}

\end{figure}

The \texttt{spotRiverGUI} automatically determines the hyperparameters
for the selected online machine learning model and adapts the input
fields to the model hyperparameters. The user can modify the
hyperparameters in the GUI. Figure~\ref{fig-spotRiverGUI-01} shows the
\texttt{spotRiverGUI} when the \texttt{forest.AMFClassifier} is selected
and Figure~\ref{fig-spotRiverGUI-02} shows the \texttt{spotRiverGUI}
when the \texttt{tree.HoeffdingTreeClassifier} is selected.

Numerical and categorical hyperparameters are treated differently in the
\texttt{spotRiverGUI}:

\begin{itemize}
\tightlist
\item
  The user can modify the lower and upper bounds for the numerical
  hyperparameters.
\item
  There are no upper or lower bounds for categorical hyperparameters.
  Instead, hyperparameter values for the categorical hyperparameters are
  considered as sets of values, e.g., the set of
  \texttt{ExhaustiveSplitter}, \texttt{HistogramSplitter},
  \texttt{GaussianSplitter} is provided for the \texttt{splitter}
  hyperparameter of the \texttt{tree.HoeffdingAdaptiveTreeClassifier}
  model as can be seen in Figure~\ref{fig-spotRiverGUI-02}. The user can
  select the full set or any subset of the set of values for the
  categorical hyperparameters.
\end{itemize}

In addition to the lower and upper bounds (or the set of values for the
categorical hyperparameters), the \texttt{spotRiverGUI} provides
information about the \texttt{Default\ values} and the
\texttt{Transformation} function. If the \texttt{Transformation}
function is set to \texttt{None}, the values of the hyperparameters are
passed to the \texttt{spot} tuner as they are. If the
\texttt{Transformation} function is set to
\texttt{transform\_power\_2\_int}, the value \(x\) is transformed to
\(2^x\) before it is passed to the \texttt{spot} tuner.

Modifications of the \texttt{Default\ values} and
\texttt{Transformation} functions values in the \texttt{spotRiverGUI}
have no effect on the hyperparameter tuning process. This is
intensional. In future versions, the user will be able to add their own
hyperparameter dictionaries to the \texttt{spotRiverGUI}, which allows
the modification of \texttt{Default\ values} and \texttt{Transformation}
functions values. Furthermore, the \texttt{spotRiverGUI} will support
more online machine learning models in future versions.

\section{Regression}\label{sec-regression}

Regression tasks will be supported soon. The same workflow as for the
binary classification task will be used, i.e., the user can select the
data set, the preprocessing model, the metric, and the online machine
learning model.

\section{Showing the Data}\label{sec-showing-data}

\begin{figure}

\centering{

\includegraphics[width=0.5\textwidth,height=\textheight]{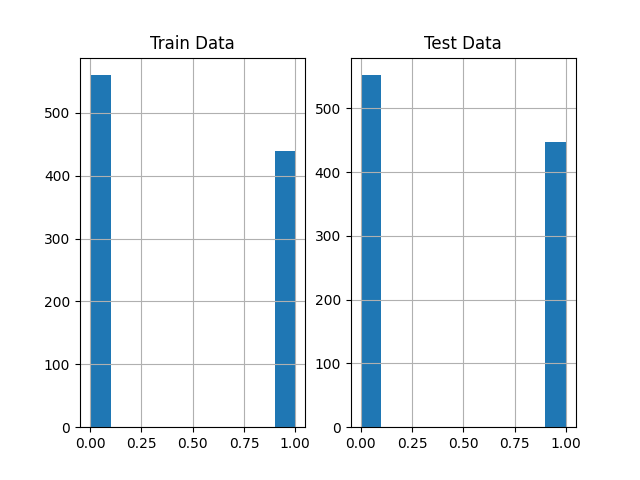}

}

\caption{\label{fig-bananas-01}Output from the \texttt{spotRiverGUI}
when \texttt{Bananas} data is selected for the \texttt{Show\ Data}
option}

\end{figure}

\begin{figure}

\centering{

\includegraphics[width=0.5\textwidth,height=\textheight]{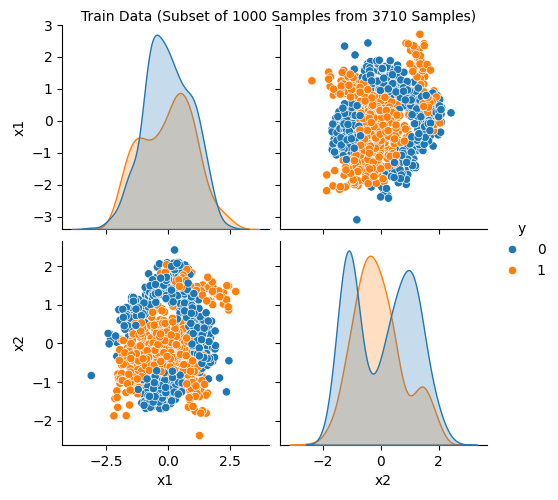}

}

\caption{\label{fig-bananas-02}Visualization of the train data. Output
from the \texttt{spotRiverGUI} when \texttt{Bananas} data is selected
for the \texttt{Show\ Data} option}

\end{figure}

\begin{figure}

\centering{

\includegraphics[width=0.5\textwidth,height=\textheight]{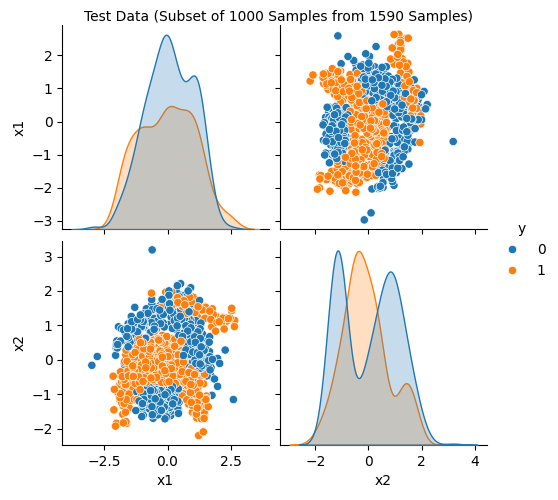}

}

\caption{\label{fig-bananas-03}Visualization of the test data. Output
from the \texttt{spotRiverGUI} when \texttt{Bananas} data is selected
for the \texttt{Show\ Data} option}

\end{figure}

The \texttt{spotRiverGUI} provides the \texttt{Show\ Data} button, which
opens a new window and shows information about the data set. The first
figure (Figure~\ref{fig-bananas-01}) shows histograms of the target
variables in the train and test data sets. The second figure
(Figure~\ref{fig-bananas-02}) shows scatter plots of the features in the
train data set. The third figure (Figure~\ref{fig-bananas-03}) shows the
corresponding scatter plots of the features in the test data set.

\begin{tcolorbox}[enhanced jigsaw, title=\textcolor{quarto-callout-note-color}{\faInfo}\hspace{0.5em}{Size of the Displayed Data Sets}, left=2mm, colframe=quarto-callout-note-color-frame, opacitybacktitle=0.6, opacityback=0, toptitle=1mm, bottomtitle=1mm, rightrule=.15mm, leftrule=.75mm, coltitle=black, breakable, toprule=.15mm, bottomrule=.15mm, colbacktitle=quarto-callout-note-color!10!white, arc=.35mm, titlerule=0mm, colback=white]

\begin{itemize}
\tightlist
\item
  Some data sets are quite large and the display of the data sets might
  take some time.
\item
  Therefore, a random subset of 1000 instances of the data set is
  displayed if the data set is larger than 1000 instances.
\end{itemize}

\end{tcolorbox}

\begin{figure}

\centering{

\includegraphics[width=0.5\textwidth,height=\textheight]{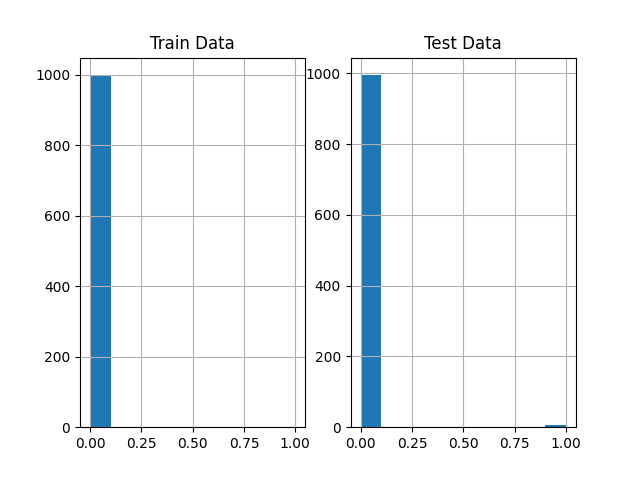}

}

\caption{\label{fig-http-01}Output from the \texttt{spotRiverGUI} when
\texttt{HTTP} data is selected for the \texttt{Show\ Data} option. The
target variable is biased.}

\end{figure}

\begin{figure}

\centering{

\includegraphics[width=0.6\textwidth,height=\textheight]{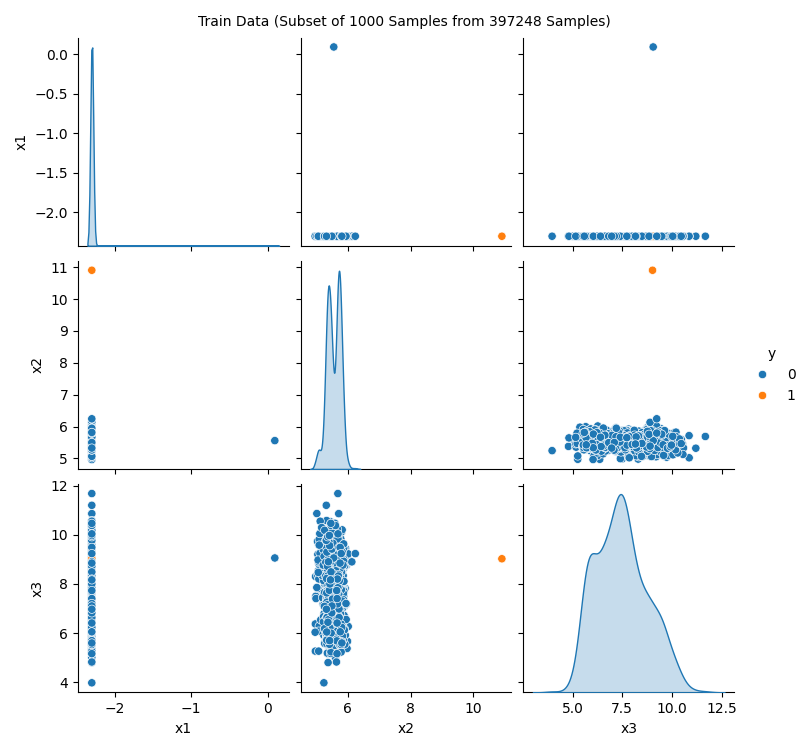}

}

\caption{\label{fig-http-02}Output from the \texttt{spotRiverGUI} when
\texttt{HTTP} data is selected for the \texttt{Show\ Data} option. A
subset of 1000 randomly chosen data points is shown. Only a few positive
events are in the data.}

\end{figure}

\begin{figure}

\centering{

\includegraphics[width=0.6\textwidth,height=\textheight]{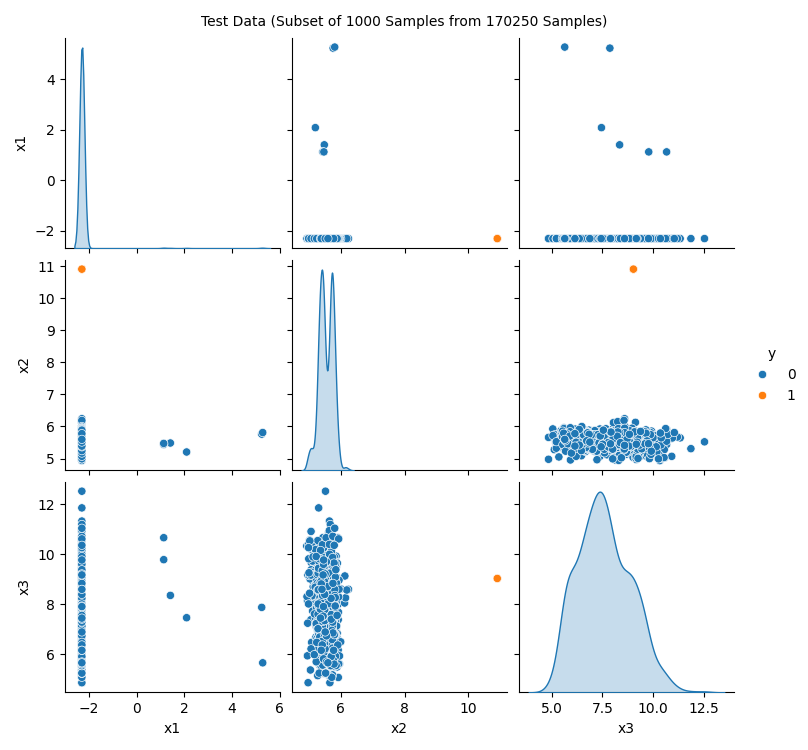}

}

\caption{\label{fig-http-03}Output from the \texttt{spotRiverGUI} when
\texttt{HTTP} data is selected for the \texttt{Show\ Data} option. The
test data set shows the same structure as the train data set.}

\end{figure}

Showing the data is important, especially for the new / unknown data
sets as can be seen in Figure~\ref{fig-http-01},
Figure~\ref{fig-http-02}, and Figure~\ref{fig-http-03}: The target
variable is highly biased. The user can check whether the data set is
correctly formatted and whether the target variable is correctly
specified.

In addition to the histograms and scatter plots, the
\texttt{spotRiverGUI} provides textual information about the data set in
the console window. e.g., for the \texttt{Bananas} data set, the
following information is shown:

\begin{Shaded}
\begin{Highlighting}[]
\NormalTok{Train data summary:}
\NormalTok{                 x1           x2            y}
\NormalTok{count  3710.000000  3710.000000  3710.000000}
\NormalTok{mean     {-}0.016243     0.002430     0.451482}
\NormalTok{std       0.995490     1.001150     0.497708}
\NormalTok{min      {-}3.089839    {-}2.385937     0.000000}
\NormalTok{25\%      {-}0.764512    {-}0.914144     0.000000}
\NormalTok{50\%      {-}0.027259    {-}0.033754     0.000000}
\NormalTok{75\%       0.745066     0.836618     1.000000}
\NormalTok{max       2.754447     2.517112     1.000000}

\NormalTok{Test data summary:}
\NormalTok{                 x1           x2            y}
\NormalTok{count  1590.000000  1590.000000  1590.000000}
\NormalTok{mean      0.037900    {-}0.005670     0.440881}
\NormalTok{std       1.009744     0.997603     0.496649}
\NormalTok{min      {-}2.980834    {-}2.199138     0.000000}
\NormalTok{25\%      {-}0.718710    {-}0.911151     0.000000}
\NormalTok{50\%       0.034858    {-}0.046502     0.000000}
\NormalTok{75\%       0.862049     0.806506     1.000000}
\NormalTok{max       2.813360     3.194302     1.000000}
\end{Highlighting}
\end{Shaded}

\section{Saving and Loading}\label{sec-saving-loading}

\subsection{Saving the Experiment}\label{saving-the-experiment}

If the experiment should not be started immediately, the user can save
the experiment by clicking on the \texttt{Save\ Experiment} button. The
\texttt{spotRiverGUI} will save the experiment as a pickle file. The
file name is generated based on the \texttt{PREFIX} parameter. The
pickle file contains a set of dictionaries, which are used to start the
experiment.

\texttt{spotRiverGUI} shows a summary of the selected hyperparameters in
the console window as can be seen in Table~\ref{tbl-hyperdict}.

\begin{longtable}[]{@{}
  >{\raggedright\arraybackslash}p{(\columnwidth - 10\tabcolsep) * \real{0.2526}}
  >{\raggedright\arraybackslash}p{(\columnwidth - 10\tabcolsep) * \real{0.0842}}
  >{\raggedright\arraybackslash}p{(\columnwidth - 10\tabcolsep) * \real{0.1895}}
  >{\raggedright\arraybackslash}p{(\columnwidth - 10\tabcolsep) * \real{0.1263}}
  >{\raggedright\arraybackslash}p{(\columnwidth - 10\tabcolsep) * \real{0.1053}}
  >{\raggedright\arraybackslash}p{(\columnwidth - 10\tabcolsep) * \real{0.2421}}@{}}
\caption{The hyperparameter values for the
\texttt{tree.HoeffdingAdaptiveTreeClassifier}
model.}\label{tbl-hyperdict}\tabularnewline
\toprule\noalign{}
\begin{minipage}[b]{\linewidth}\raggedright
name
\end{minipage} & \begin{minipage}[b]{\linewidth}\raggedright
type
\end{minipage} & \begin{minipage}[b]{\linewidth}\raggedright
default
\end{minipage} & \begin{minipage}[b]{\linewidth}\raggedright
lower
\end{minipage} & \begin{minipage}[b]{\linewidth}\raggedright
upper
\end{minipage} & \begin{minipage}[b]{\linewidth}\raggedright
transform
\end{minipage} \\
\midrule\noalign{}
\endfirsthead
\toprule\noalign{}
\begin{minipage}[b]{\linewidth}\raggedright
name
\end{minipage} & \begin{minipage}[b]{\linewidth}\raggedright
type
\end{minipage} & \begin{minipage}[b]{\linewidth}\raggedright
default
\end{minipage} & \begin{minipage}[b]{\linewidth}\raggedright
lower
\end{minipage} & \begin{minipage}[b]{\linewidth}\raggedright
upper
\end{minipage} & \begin{minipage}[b]{\linewidth}\raggedright
transform
\end{minipage} \\
\midrule\noalign{}
\endhead
\bottomrule\noalign{}
\endlastfoot
grace\_period & int & 200 & 10 & 1000 & None \\
max\_depth & int & 20 & 2 & 20 & transform\_power\_2\_int \\
delta & float & 1e-07 & 1e-08 & 1e-06 & None \\
tau & float & 0.05 & 0.01 & 0.1 & None \\
leaf\_prediction & factor & nba & 0 & 2 & None \\
nb\_threshold & int & 0 & 0 & 10 & None \\
splitter & factor & GaussianSplitter & 0 & 2 & None \\
bootstrap\_sampling & factor & 0 & 0 & 1 & None \\
drift\_window\_threshold & int & 300 & 100 & 500 & None \\
drift\_detector & factor & ADWIN & 0 & 0 & None \\
switch\_significance & float & 0.05 & 0.01 & 0.1 & None \\
binary\_split & factor & 0 & 0 & 1 & None \\
max\_size & float & 100.0 & 100 & 1000 & None \\
memory\_estimate\_period & int & 1000000 & 100000 & 1e+06 & None \\
stop\_mem\_management & factor & 0 & 0 & 1 & None \\
remove\_poor\_attrs & factor & 0 & 0 & 1 & None \\
merit\_preprune & factor & 0 & 0 & 1 & None \\
\end{longtable}

\subsection{Loading an Experiment}\label{loading-an-experiment}

Future versions of the \texttt{spotRiverGUI} will support the loading of
experiments from the GUI. Currently, the user can load the experiment by
executing the command \texttt{load\_experiment}, see
\url{https://sequential-parameter-optimization.github.io/spotPython/reference/spotPython/utils/file/\#spotPython.utils.file.load_experiment}.

\section{Running a New Experiment}\label{sec-running-experiment}

An experiment can be started by clicking on the \texttt{Run\ Experiment}
button. The GUI calls \texttt{run\_spot\_python\_experiment} from
\texttt{spotGUI.tuner.spotRun}. Output will be shown in the console
window from which the GUI was started.

\subsection{Starting and Stopping
Tensorboard}\label{starting-and-stopping-tensorboard}

\begin{figure}

\centering{

\includegraphics[width=1\textwidth,height=\textheight]{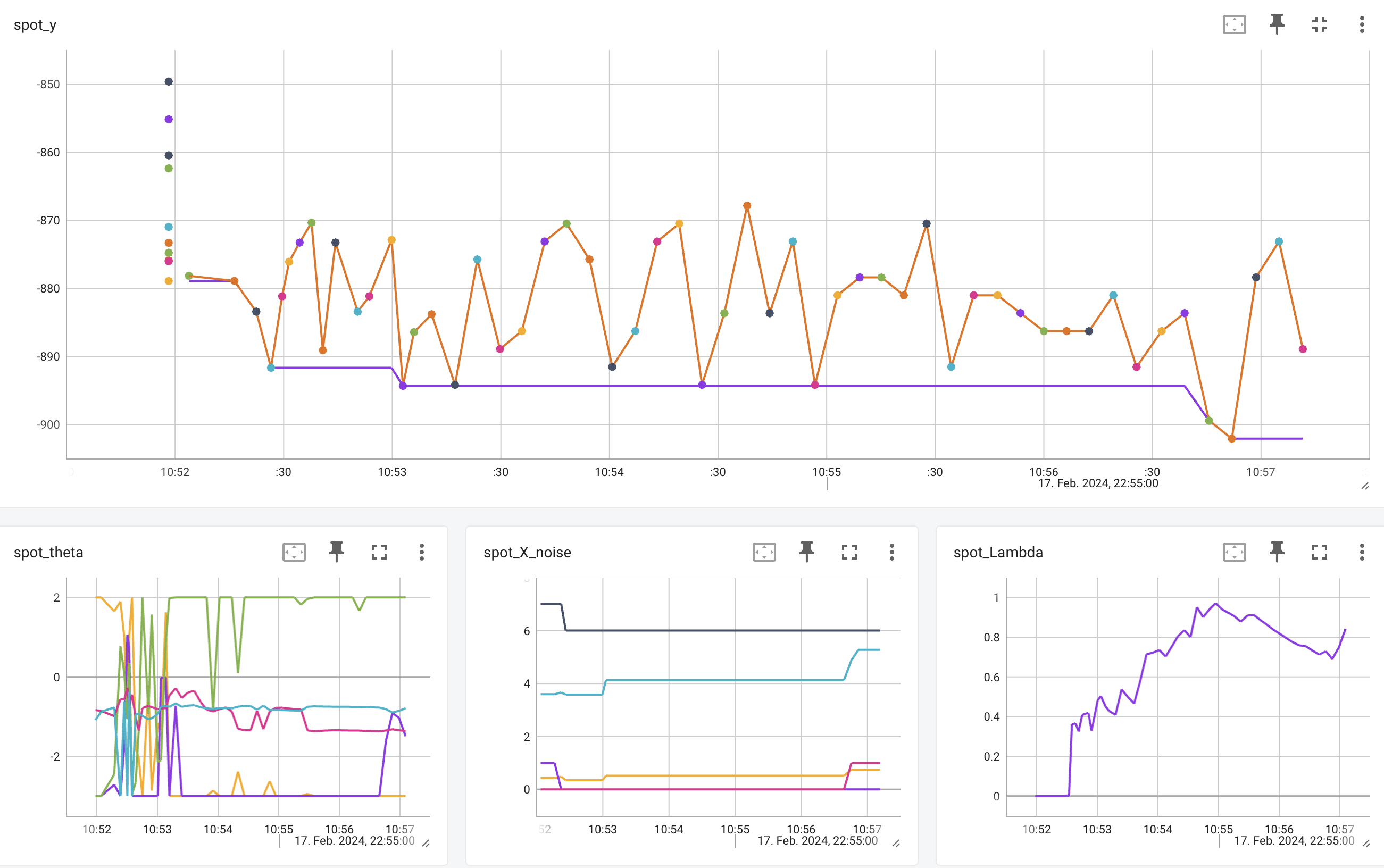}

}

\caption{\label{fig-tensorboard-05}Tensorboard visualization of the
hyperparameter tuning process}

\end{figure}

\begin{figure}

\centering{

\includegraphics[width=0.7\textwidth,height=\textheight]{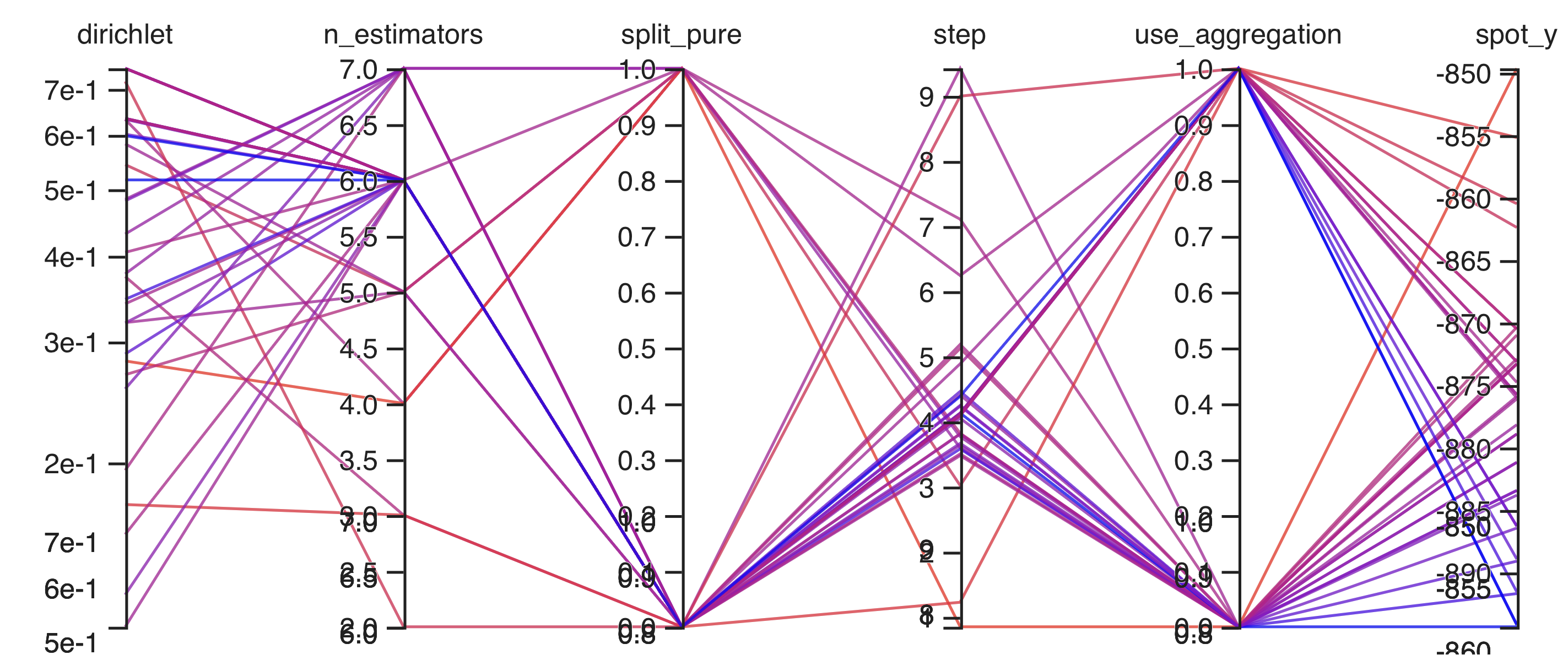}

}

\caption{\label{fig-tensorboard-04}Tensorboard. Parallel coordinates
plot}

\end{figure}

Tensorboard (Abadi et al. 2016) is automatically started when an
experiment is started. The tensorboard process can be observed in a
browser by opening the \url{http://localhost:6006} page. Tensorboard
provides a visual representation of the hyperparameter tuning process.
Figure~\ref{fig-tensorboard-05} and Figure~\ref{fig-tensorboard-04} show
the tensorboard page when the \texttt{spotRiverGUI} is performing the
tuning process.

\texttt{spotPython.utils.tensorboard} provides the methods
\texttt{start\_tensorboard} and \texttt{stop\_tensorboard} to start and
stop tensorboard as a background process. After the experiment is
finished, the tensorboard process is stopped automatically.

\section{Performing the Analysis}\label{sec-analysis}

If the hyperparameter tuning process is finished, the user can analyze
the results by clicking on the \texttt{Analysis} button. The following
options are available:

\begin{itemize}
\tightlist
\item
  Progress plot
\item
  Compare tuned versus default hyperparameters
\item
  Importance of hyperparameters
\item
  Contour plot
\item
  Parallel coordinates plot
\end{itemize}

\begin{figure}

\centering{

\includegraphics[width=0.7\textwidth,height=\textheight]{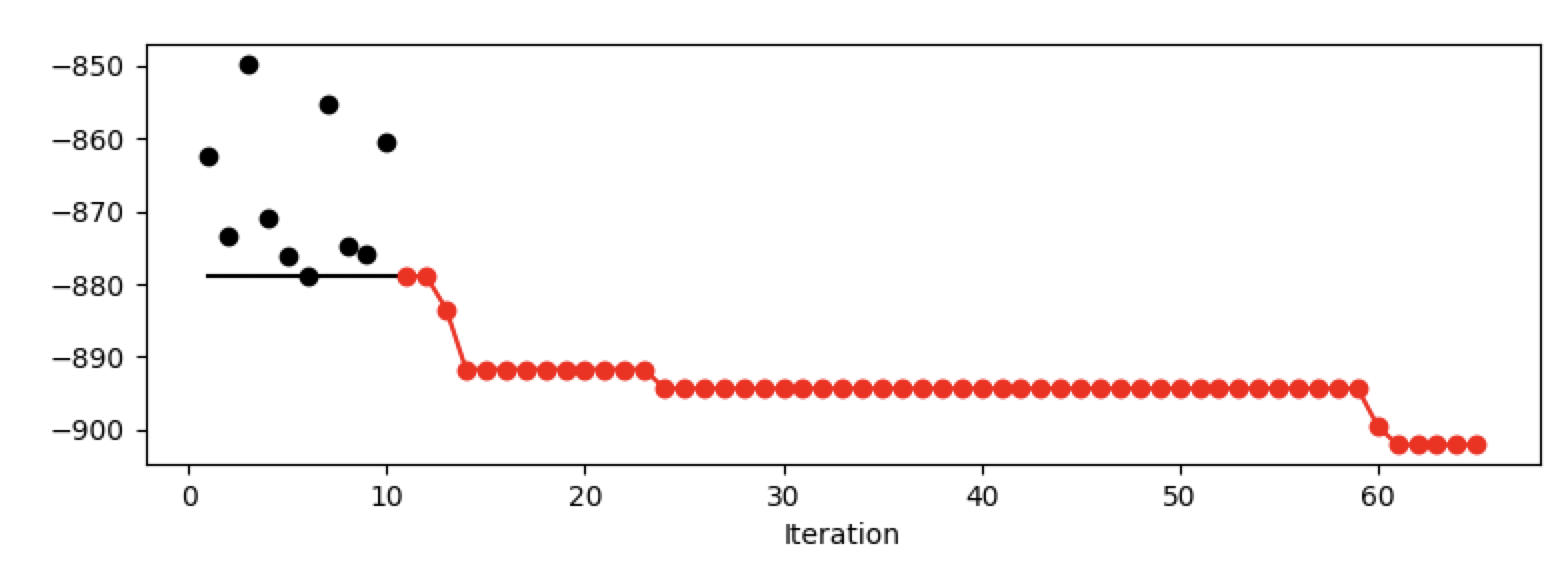}

}

\caption{\label{fig-prg-00}Progress plot of the hyperparameter tuning
process}

\end{figure}

\begin{figure}

\centering{

\includegraphics[width=0.8\textwidth,height=\textheight]{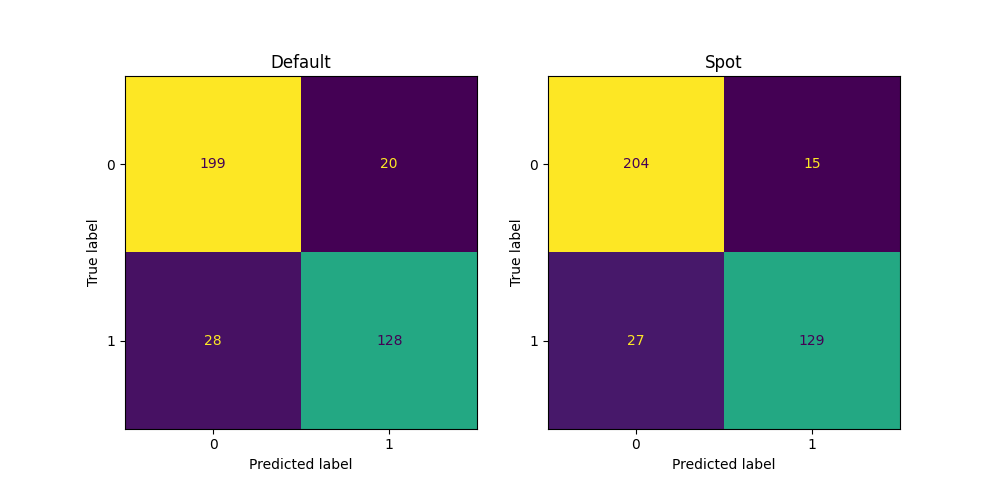}

}

\caption{\label{fig-cm-00}Confusion matrix}

\end{figure}

\begin{figure}

\centering{

\includegraphics{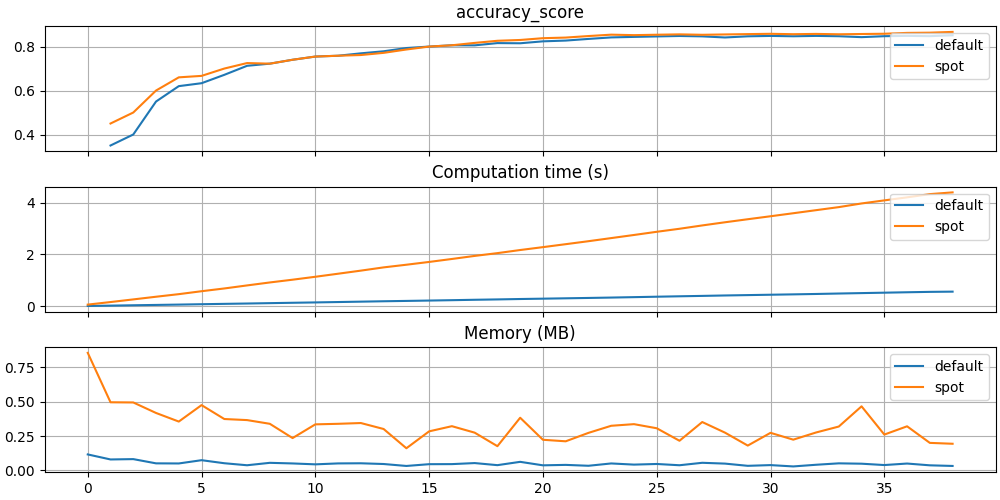}

}

\caption{\label{fig-default-tuned-00}Default versus tuned
hyperparameters}

\end{figure}

\begin{figure}

\centering{

\includegraphics[width=0.7\textwidth,height=\textheight]{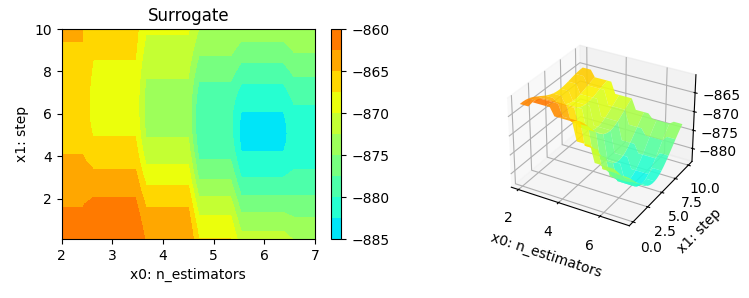}

}

\caption{\label{fig-surrogate-00}Surrogate plot based on the Kriging
model. \texttt{x0} and \texttt{x1} plotted against each other.}

\end{figure}

\begin{figure}

\centering{

\includegraphics[width=0.7\textwidth,height=\textheight]{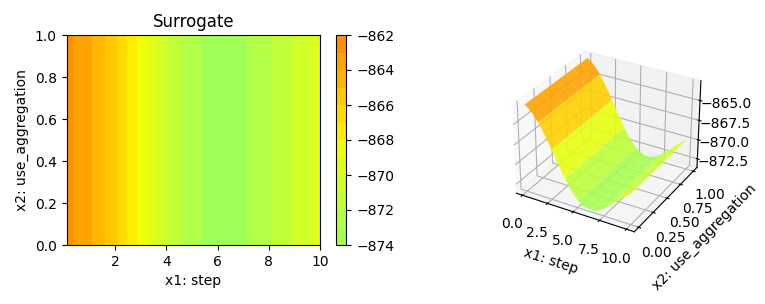}

}

\caption{\label{fig-surrogate-01}Surrogate plot based on the Kriging
model. \texttt{x1} and \texttt{x2} plotted against each other.}

\end{figure}

\begin{figure}

\centering{

\includegraphics[width=0.7\textwidth,height=\textheight]{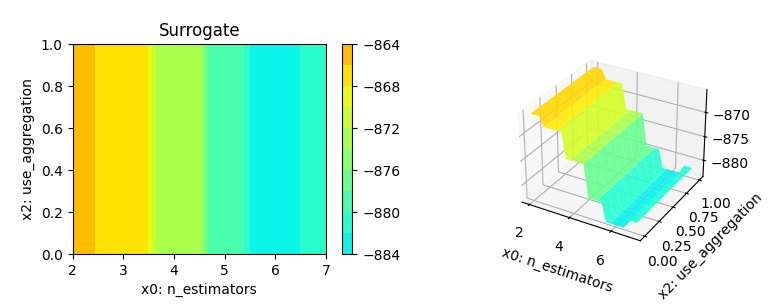}

}

\caption{\label{fig-surrogate-02}Surrogate plot based an the Kriging
model. \texttt{x0} and \texttt{x2} plotted against each other.}

\end{figure}

Figure~\ref{fig-prg-00} shows the progress plot of the hyperparameter
tuning process. Black dots denote results from the initial design. Red
dots illustrate the improvement found by the surrogate model based
optimization. For binary classification tasks, the
\texttt{roc\_auc\_score} can be used as the evaluation metric. The
confusion matrix is shown in Figure~\ref{fig-cm-00}. The default versus
tuned hyperparameters are shown in Figure~\ref{fig-default-tuned-00}.
The surrogate plot is shown in Figure~\ref{fig-surrogate-00},
Figure~\ref{fig-surrogate-01}, and Figure~\ref{fig-surrogate-02}.

Furthermore, the tuned hyperparameters are shown in the console window.
A typical output is shown below (modified due to formatting reasons):

\begin{Shaded}
\begin{Highlighting}[]
\NormalTok{|name    |type   |default |low | up |tuned |transf |importance|stars|}
\NormalTok{|{-}{-}{-}{-}{-}{-}{-}{-}|{-}{-}{-}{-}{-}{-}{-}|{-}{-}{-}{-}{-}{-}{-}{-}|{-}{-}{-}{-}|{-}{-}{-}{-}|{-}{-}{-}{-}{-}{-}|{-}{-}{-}{-}{-}{-}{-}|{-}{-}{-}{-}{-}{-}{-}{-}{-}{-}|{-}{-}{-}{-}{-}|}
\NormalTok{|n\_estim |int    |    3.0 |2.0 |7.0 |  3.0 | pow\_2 |      0.04|     |}
\NormalTok{|step    |float  |    1.0 |0.1 |10.0|  5.12| None  |      0.21| .   |}
\NormalTok{|use\_agg |factor |    1.0 |0.0 |1.0 |  0.0 | None  |     10.17| *   |}
\NormalTok{|dirichl |float  |    0.5 |0.1 |0.75|  0.37| None  |     13.64| *   |}
\NormalTok{|split\_p |factor |    0.0 |0.0 |1.0 |  0.0 | None  |    100.00| *** |}
\end{Highlighting}
\end{Shaded}

In addition to the tuned parameters that are shown in the column
\texttt{tuned}, the columns \texttt{importance} and \texttt{stars} are
shown. Both columns show the most important hyperparameters based on
information from the surrogate model. The \texttt{stars} column shows
the importance of the hyperparameters in a graphical way. It is
important to note that the results are based on a demo of the
hyperparameter tuning process. The plots are not based on a real
hyperparameter tuning process. The reader is referred to
Bartz-Beielstein (2024b) for further information about the analysis of
the hyperparameter tuning process.

\section{Summary and Outlook}\label{sec-summary}

The \texttt{spotRiverGUI} provides a graphical user interface for the
\texttt{spotRiver} package. It releases the user from the burden of
manually searching for the optimal hyperparameter setting. After copying
a data set into the \texttt{userData} folder and starting
\texttt{spotRiverGUI}, users can compare different OML algorithms from
the powerful \texttt{river} package in a convenient way. Users can
generate configurations on their local machines, which can be
transferred to a remote machine for execution. Results from the remote
machine can be copied back to the local machine for analysis.

\begin{tcolorbox}[enhanced jigsaw, title=\textcolor{quarto-callout-important-color}{\faExclamation}\hspace{0.5em}{Benefits of the spotRiverGUI:}, left=2mm, colframe=quarto-callout-important-color-frame, opacitybacktitle=0.6, opacityback=0, toptitle=1mm, bottomtitle=1mm, rightrule=.15mm, leftrule=.75mm, coltitle=black, breakable, toprule=.15mm, bottomrule=.15mm, colbacktitle=quarto-callout-important-color!10!white, arc=.35mm, titlerule=0mm, colback=white]

\begin{itemize}
\tightlist
\item
  Very easy to use (only the data must be provided in the correct
  format).
\item
  Reproducible results.
\item
  State-of-the-art hyperparameter tuning methods.
\item
  Powerful analysis tools, e.g., Bayesian optimization (Forrester,
  Sóbester, and Keane 2008; Gramacy 2020).
\item
  Visual representation of the hyperparameter tuning process with
  tensorboard.
\item
  Most advanced online machine learning models from the \texttt{river}
  package.
\end{itemize}

\end{tcolorbox}

The \texttt{river} package (Montiel et al. 2021), which is very well
documented, can be downloaded from \url{https://riverml.xyz/latest/}.

The \texttt{spotRiverGUI} is under active development and new features
will be added soon. It can be downloaded from GitHub:
\url{https://github.com/sequential-parameter-optimization/spotGUI}.

Interactive Jupyter Notebooks and further material about OML are
provided in the GitHub repository
\url{https://github.com/sn-code-inside/online-machine-learning}. This
material is part of the supplementary material of the book ``Online
Machine Learning - A Practical Guide with Examples in Python'', see
\url{https://link.springer.com/book/9789819970063} and the forthcoming
book ``Online Machine Learning - Eine praxisorientierte Einführung'',
see \url{https://link.springer.com/book/9783658425043}.

\newpage{}

\section*{References}\label{references}
\addcontentsline{toc}{section}{References}

\phantomsection\label{refs}

\end{document}